\documentclass[twoside]{article}

\usepackage{amsmath}
\usepackage{amssymb}
\usepackage{amsthm}
\usepackage{bbm}
\usepackage{natbib}
\usepackage{graphicx}
\usepackage{cleveref}
\usepackage{todonotes}
\usepackage{subfigure}
\usepackage{multirow}
\usepackage{cases}
\usepackage{booktabs}

\newtheorem{prop}{Proposition}
\newtheorem{assum}{Assumption}
\newtheorem{lemma}{Lemma}
\newtheorem{rem}{Remark}
\newtheorem{theorem}{Theorem}

\usepackage[accepted]{arxiv}
\begin{document}

%

%

\twocolumn[

\aistatstitle{Finite Sample Confidence Regions for Linear Regression Parameters Using Arbitrary Predictors}

\aistatsauthor{ Charles Guille-Escuret \And Eugene Ndiaye }

\aistatsaddress{ Apple \And  Apple } ]

\begin{abstract}
  We explore a novel methodology for constructing confidence regions for parameters of linear models, using predictions from any arbitrary predictor. Our framework requires minimal assumptions on the noise and can be extended to functions deviating from strict linearity up to some adjustable threshold, thereby accommodating a comprehensive and pragmatically relevant set of functions. The derived confidence regions can be cast as constraints within a Mixed Integer Linear Programming framework, enabling optimisation of linear objectives. This representation enables robust optimization and the extraction of confidence intervals for specific parameter coordinates. Unlike previous methods, the confidence region can be empty, which can be used for hypothesis testing. Finally, we validate the empirical applicability of our method on synthetic data.
\end{abstract}

\section{Introduction}
\label{sec:intro}

Estimating the parameters of an unknown linear system using noisy observations stands as a cornerstone challenge in various disciplines including signal processing, system identification, control theory, and statistics. Mostly, the current methods yield point estimates. To incorporate the inherent uncertainty associated with these estimated parameters, one can delineate confidence regions. Such regions ensure, with a high probability, that the true parameter resides within them. Confidence regions are crucial when robustness is a priority, offering direct utility in both uncertainty quantification and robust optimization.

Historically in statistics, confidence regions are predominantly derived using closed-form solutions, often predicated on the assumption of constant additive Gaussian noise \citep{draper1981applied}. Such an assumption curtails their practical utility in real-world scenarios, where heteroscedasticity might prevail and noise may manifest in intricate functional forms. More contemporary techniques promise confidence regions with finite sample coverage guarantees even under considerably relaxed noise assumptions. Nevertheless, these methods often limit themselves to membership testing (i.e., ascertaining if a particular parameter falls within the confidence region), without offering a compact representation \citep{CAMPI20051751, DENDEKKER20085024}. This characteristic hinders their applicability to robust optimization and uncertainty quantification. 

In this study, we introduce Residual Intervals Inversion (RII), a novel methodology for the construction of confidence regions pertaining to linear model parameters. Central to our approach is the harnessing of predictions from an arbitrary, ad hoc predictor. Such a predictor might be sourced from conventional tools like the Least Square (LS) estimator or more complex non-linear models.

Our only assumption on the noise is that it must possess a median of zero across the entire input space. This is much weaker than the assumption of Gaussian noise made in statistical approaches, and even weaker than symmetric noise assumptions made in more recent research \citep{Csaji_2015, MSPS, CAMPI20051751}. Additionally, our approach integrates an adjustable tolerance parameter that can relax this condition by bounding the noise's quantile deviation, thereby granting additional flexibility.

The confidence region is represented by a set of linear inequalities in the model parameter and binary variables controlling which inequalities are active. This formulation seamlessly permits its representation as constraints within a Mixed-Integer Linear Programming (MILP) problem. As a result, linear or quadratic objectives can be optimized over these confidence regions, enabling tasks such as computing confidence intervals for specific parameter coordinates.

Notably, when the ad hoc predictor substantially outperforms any linear counterpart, the confidence regions we construct may be empty. This may occur either for non-linear ad-hoc predictors, or when it has access to different input variables. In contrast to previous works, our method thus exhibits the capacity to reject the null hypothesis, signaling that the data might not exhibit a linear relationship with the specified input. This capability paves the way for its use in hypothesis testing and feature selection.

The most salient properties of our method can be summarized as follows:
\begin{itemize}
    \item Capability to use strong predictors (including non-linear) to obtain smaller confidence regions.
    \item The noise is only assumed to have a median value of zero everywhere. This assumption can be flexibly relaxed by introducing user-specified tolerance level.\looseness=-1
    \item The possibility to optimize linear and quadratic objectives over the confidence region by solving a MILP or MIQP problem.
    \item The confidence regions ensure finite-sample validity and for any user-determined target coverage.
\end{itemize}


\section{Related Work}
\label{sec:related_work}

Estimating the parameters of dynamical systems is a cornerstone challenge of system identification \citep{4019326, söderström1989system, LJUNG20101, ljung1994modeling, ljung1999system}. The LS estimator error is asymptotically normal under reasonable assumptions, which can be used to derive confidence regions. Recently, \citet{doi:10.1080/01621459.2020.1831924} proposed robust estimators in heteroskedastic settings. However, these methods are only asymptotically valid and provide no guarantees in practical settings where available data is finite. 

Other methods have derived finite sample valid confidence regions. \citet{doi:10.1073/pnas.1922664117} relies on computing the likelihood which is typically difficult in the presence of unknown and non-standard noise. \citet{10.1214/aoms/1177728718} is distribution-free but constructs unbounded confidence regions, which is impractical.

Other contemporary works have focused on methods to construct finite-sample valid confidence regions with weak assumptions on the noise \citep{CAMPI20051751, DALAI20071418, DENDEKKER20085024}, but only provide a method to infer whether a given parameter $\theta$ belongs in the confidence region (membership testing), without compact formulation, hence limiting downstream applications.

Perhaps the closest work in the literature is SPS \citep{Csaji_2015,CSAJI2012227} which constructs finite sample valid confidence regions for any symmetric noise, in the form of an ellipsoid centered on the LS estimator. Similarly to RII, linear and quadratic objectives can thus be optimized over the confidence regions. We compare RII to SPS in section \ref{sec:experiments}.

Among other relevant works, \citet{dobriban2023joint} derives joint confidence regions over the prediction and parameters, and \citet{angelopoulos2023prediction} uses an ad hoc predictor and unlabeled data to infer confidence sets over various statistics, including linear parameters, but only guarantee asymptotic validity (non-asymptotic results are obtained under stronger assumption on the distribution e.g. bounded support).

\section{Problem Setting}
\label{sec:problem_setting}

In this section we introduce the notations and assumptions that will be used throughout this work. 

For $m\in\mathbb{N}$, let $[m]$ denote the set $\{1, \dots, m\}$.

Consider the following linear regression system
\begin{equation}
    \label{eq:linear_model}
    Y = \theta_\star^\top X + \varepsilon,
\end{equation}
where $Y \in \mathbb{R}$ is the target variable, $\theta_\star \in \mathbb{R}^d$ is the ground truth parameter to be estimated, $X \in \mathbb{R}^d$ the input variable, and $\varepsilon \in \mathbb{R}$ is the noise.

We consider a finite sample of size $n$ which consists of \texttt{inputs} $$\pmb{X}=X_1, \dots, X_n, $$ \texttt{noise} $$\pmb{\varepsilon}=\varepsilon_1, \dots, \varepsilon_n$$ and \texttt{targets} $$\pmb{Y}=Y_1, \dots, Y_n.$$ 

\subsection{Assumptions}

Since homoscedasticity is rarely verified in real world systems, we allow the noise $\varepsilon$ to depend on $X$ (heteroscedasticity). Our first assumption is the independence of the noise, conditionally on $X$:
 
\begin{assum}[Conditional independence]\label{assum:cond_inde} We suppose that the noise is conditionally independent, given inputs
$$ \forall i\neq j, (\varepsilon_i \perp \!\!\! \perp \varepsilon_j \mid X_i) $$
\end{assum}

When the noise does not depend on $X$ we recover the assumption of independent noise that is standard in the literature. 

Without any further assumption on $\varepsilon$, any arbitrary function $f$ of $X$ would verify \Cref{eq:linear_model} for any $\theta_\star$ with $$\varepsilon=f(X)-\theta_\star^T X.$$ For our model to be informative, it is therefore necessary to adopt some restrictions on the noise. 

Recent works have departed from the usual assumption of normally distributed noise, to make confidence regions more applicable to realistic settings.

We introduce a tolerance parameter $b$ such that
$$0\leq b\leq \frac{1}{2}$$
which controls how strict our assumption on the noise is.
Even when $b=0.5$, \eqref{eq:b-bounded_quantile-1} and \eqref{eq:b-bounded_quantile-2} are equivalent to having a noise of median $0$, which is weaker than the assumption of symmetric distribution in \citep{Csaji_2015, CAMPI20051751}.

We also define
$$d_\varepsilon(X):= \min\bigg(\mathbb{P}(\varepsilon\geq 0 \mid X),\, \mathbb{P}(\varepsilon\leq 0 \mid X)\bigg).$$
We consider two versions of our second assumption, contingent on the independence of inputs.

\begin{assum}\label{assum:A_a}
When the input data $X_1,\dots,X_n$ are independent and identically distributed, we suppose
\begin{equation} 
    \label{eq:b-bounded_quantile-1}
    \mathbb{E}_X[d_\varepsilon(X)] \geq b.
\end{equation}
\end{assum}

\begin{assum}\label{assum:A_b} We suppose that
\begin{equation}
    \label{eq:b-bounded_quantile-2}
    \min_{x\in\mathbb{R}^d}(d_\varepsilon(x)) \geq b.
\end{equation}
\end{assum}

Intuitively, \Cref{assum:A_a} and \Cref{assum:A_b} ensure that $\varepsilon$ is not too likely to be positive or too likely to be negative. \Cref{assum:A_a} and \Cref{assum:A_b} lead to the same guarantees, but when the inputs are iid, \Cref{assum:A_a} is less restrictive. Unless specified otherwise, we assume that either \Cref{assum:A_a} or \Cref{assum:A_b} are verified.\looseness=-1

The assumption is strongest for $b=0.5$. As $b$ decreases, $\theta_\star^T X$ is allowed to deviate (in terms of quantile) from the median of $Y$, and the model becomes less restrictive. When $b=0$, Equations \eqref{eq:b-bounded_quantile-1} and \eqref{eq:b-bounded_quantile-2} become vacuously true statements, and the model of \Cref{eq:linear_model} describes any stochastic function of $X$ for any $\theta_\star$.

\Cref{assum:A_a} and \Cref{assum:A_b} are stable when multiplying the noise by any deterministic function of $X$. This allows for instance the seamless integration of multiplicative noise, constant by part noise, etc.

\subsection{Objectives}

\begin{figure}[ht]
    \centering
    \includegraphics[width=\linewidth]{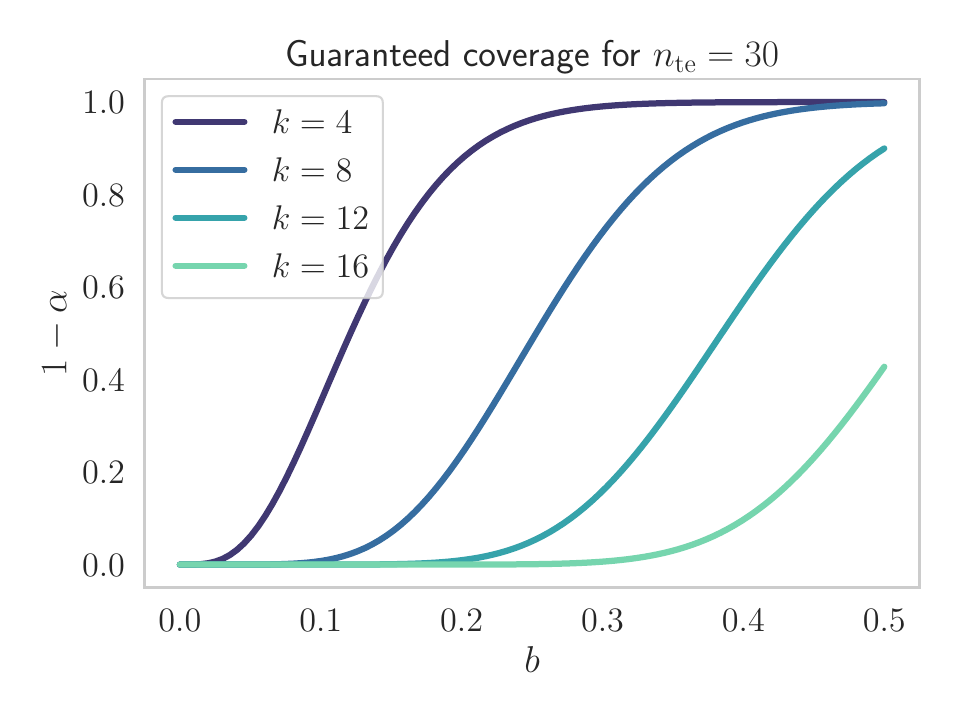}
    \caption{Guaranteed coverage $1-\alpha=S_{n_{\rm te}}(k,b)$ from \Cref{eq:coverage_guarantee} for $n_{te}=30$ and $k\in[4,8,12,16]$. 
    }
    \label{fig:coverage_by_b}
\end{figure}

Our goal is to build a confidence region $\Theta_{\alpha}(\pmb{X,Y})\subset \mathbb{R}^d$ over the unknown parameter $\theta_\star$ that has a finite sample valid coverage
\begin{equation}
    \label{eq:confidence_regions}
    \mathbb{P}_{\pmb{X,Y}}\big(\theta_\star\in \Theta_{\alpha}(\pmb{X,Y})\big)\geq 1-\alpha,
\end{equation}
For some user-specified confidence level $\alpha \in (0,1)$. While the whole output space $\mathbb{R}^d$ is a trivial solution, we aim at finding smaller sets yielding more informative results for the applications listed in \Cref{sec:applications}.
\section{Construction of Confidence Regions}
\label{sec:construction}

\begin{figure*}[t]
  \centering
  \subfigure[1D example for a linear distribution with additive Gaussian noise, using the least square linear predictor on $X$. \label{fig:1d_example_linear}]{\includegraphics[width=0.49\linewidth]{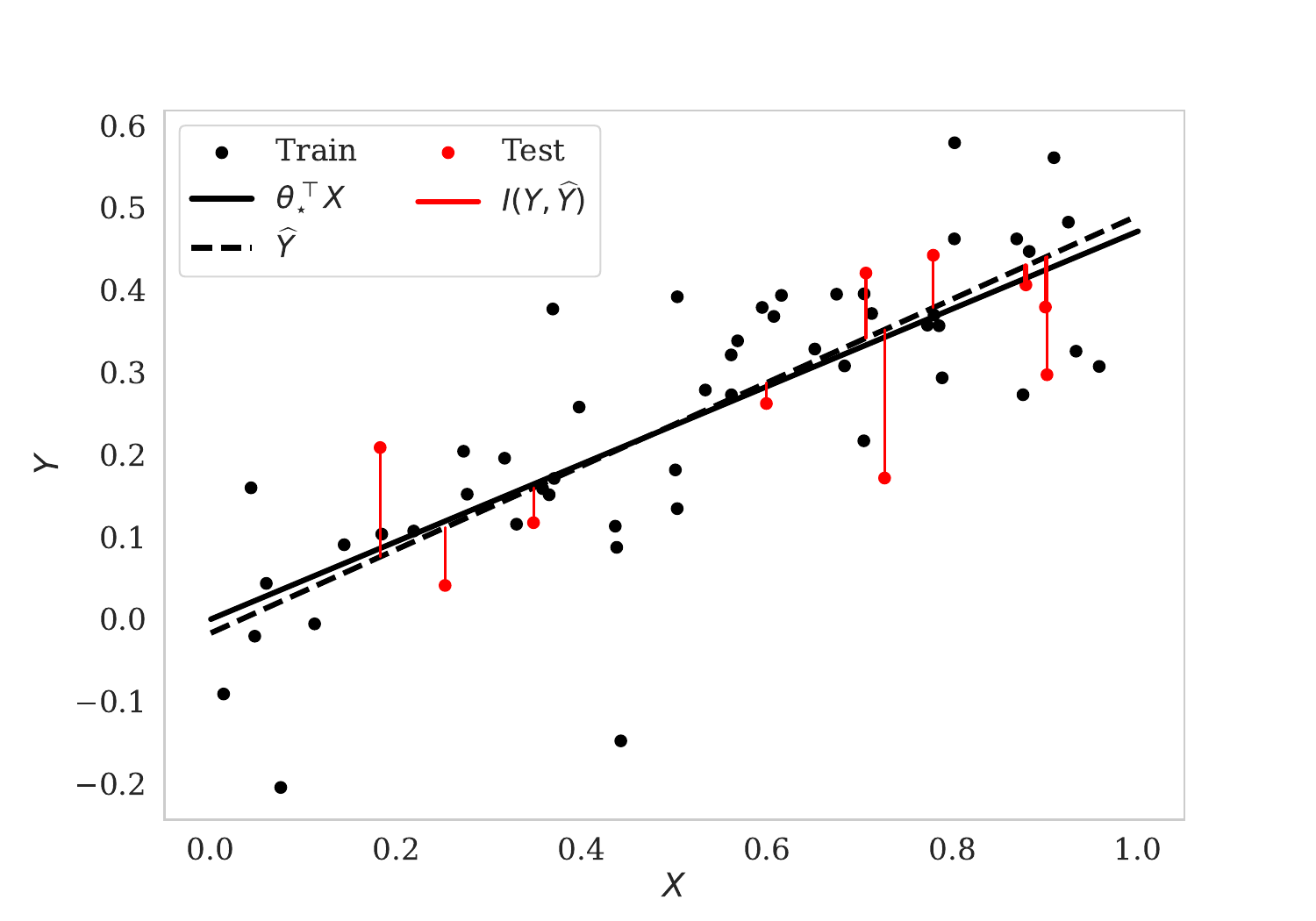}}
  \subfigure[1D example for a distribution linear in $Z=(X,sin(8\pi||X||_2))$ with additive Gaussian noise using the least square predictor on $Z$.]{\includegraphics[width=0.49\linewidth]{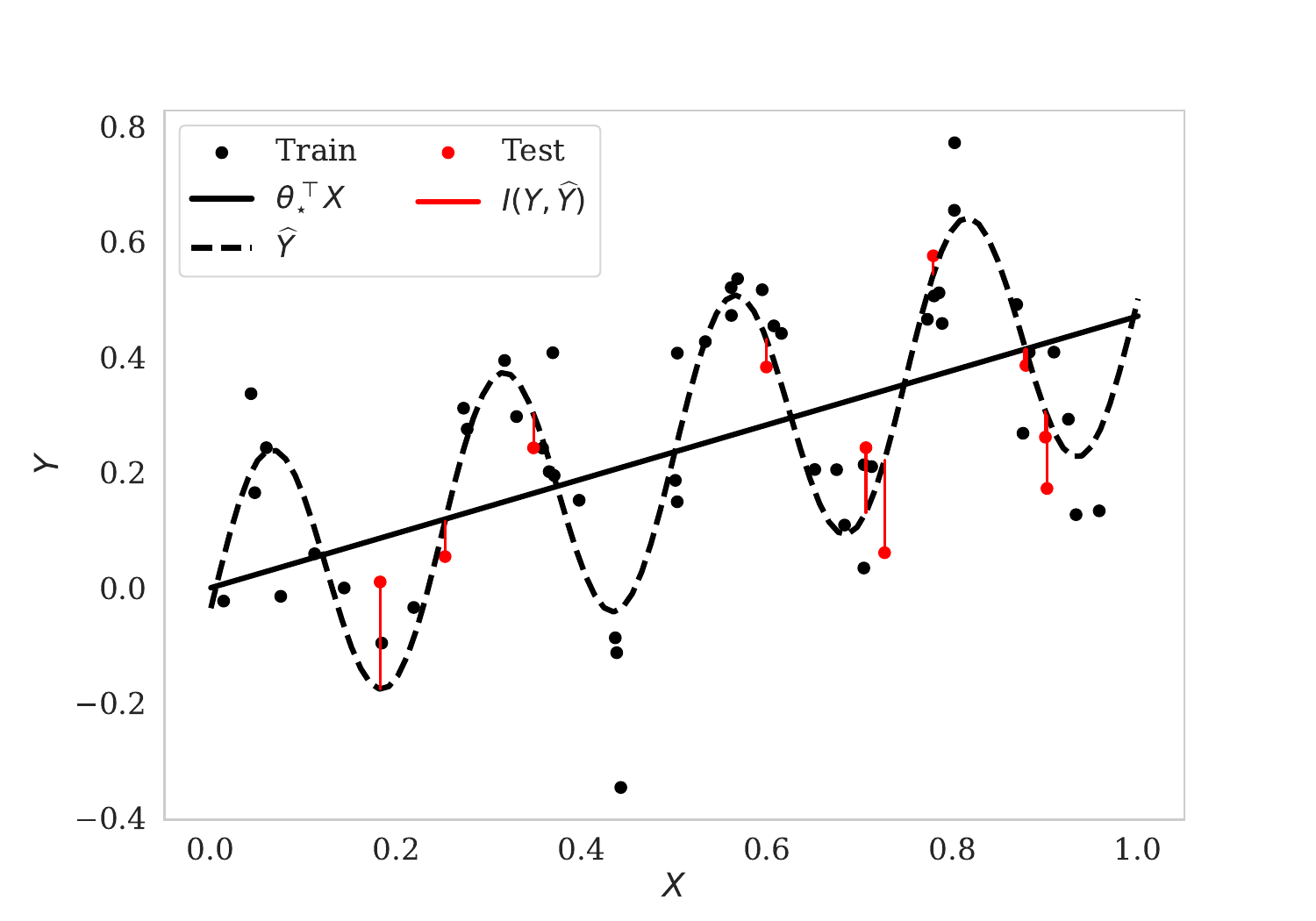}}
\caption{Illustration of residual intervals on synthetic datasets with both linear and non-linear dependence between input $X$ and output $Y$. \label{fig:illustration}}
\end{figure*}

Let $n_{te}\leq n$ and $(\pmb{X_{te},Y_{te}})$ a subset of $(\pmb{X,Y})$ of size $n_{te}$. For the sake of simplicity, let us consider the testing data \begin{align*}
\pmb{X_{te}} &=X_1,\dots,X_{n_{te}}, \\
\pmb{Y_{te}} &=Y_1,\dots,Y_{n_{te}}.
\end{align*}
Let us also denote
$$\pmb{\widehat{Y}_{{\rm te}}}=\widehat{Y}_1,\dots,\widehat{Y}_{n_{te}}$$ the predictions of $\pmb{Y_{te}}$ made by \textit{any} ad hoc predictor. The only constraint is that the predictor must not have seen the true targets $\pmb{Y_{te}}$, ie.
\begin{equation} 
\label{eq:indep_pred_test}
(\pmb{Y_{te}}\perp \!\!\! \perp \pmb{\widehat{Y}_{{\rm te}}}\mid \pmb{X_{te}}).
\end{equation}
$\pmb{\widehat{Y}_{{\rm te}}}$ can typically be the predictions of the ordinary least square model trained on the remaining samples 
\begin{align*}
\pmb{X_{tr}}&=X_{n_{te}+1},\dots, X_n, \\
\pmb{Y_{tr}}&=Y_{n_{te}+1},\dots, Y_n.
\end{align*}
Alternatively, $\pmb{\widehat{Y}_{{\rm te}}}$ can also be predictions induced by a non-linear model, a model of a different input variable $Z$, or a model trained on independent data.

Our method, RII, proceeds in two steps:
\begin{enumerate}
    \item The first step is to build intervals $I(Y,\widehat{Y})$, which we call residual intervals, such that
    $$\forall i\in [n_{\rm te}], \,\mathbb{P}_{X_i,Y_i}\left(\theta_{\star}^{\top} X_i \in I(Y_i,\widehat{Y_i}) \right)\geq b.$$
    
    \item Then, we consider as confidence region the set of $\theta \in \mathbb{R}^d$ such that $\theta^\top X \in I(Y,\widehat{Y})$ reasonably frequently in the test set, and represent it as the feasible set of a MILP.
\end{enumerate}

\subsection{Building Residual Intervals}

The first step consists in building reasonably sized intervals $I(Y,\widehat{Y})$ for each test point, so that $\theta_{\star}^{\top} X$ belongs in the interval with a guaranteed probability. 
We accomplish this by taking the interval between the true label and predicted value
\begin{equation}
    I(Y,\widehat{Y}) = [\min(Y,\widehat{Y}), \max(Y, \widehat{Y})].
\end{equation}
The intuition is that if $\widehat{Y}\geq \theta_{\star}^{\top} X$, then there is a probability at least $d_\varepsilon(X)$ that $\varepsilon\leq 0$ and thus that $$Y\leq \theta_{\star}^{\top} X \leq \widehat{Y}.$$ A similar reasoning holds when $\widehat{Y}\leq \theta_{\star}^{\top} X$.

\Cref{prop:conformal_window} formalizes this intuition and shows that this interval has a guaranteed coverage of $b$ for $\theta_{\star}^{\top} X$. A formal proof is provided in the appendix.

\begin{lemma}
\label{prop:conformal_window}
Under \Cref{assum:cond_inde} and (\Cref{assum:A_a} or \Cref{assum:A_b}), for any test point $(X, Y)$ with prediction $\widehat{Y}$, it holds
\begin{equation}
\label{eq:conformal_window}
\mathbb{P}_{X, Y}\bigg(\theta_{\star}^{\top} X\in I(Y,\widehat{Y})\bigg) \geq b .
\end{equation}
\end{lemma}

\subsection{Building the Confidence Region}

Let $\theta\in\mathbb{R}^d$. For $i\in [n_{\rm te}]$, we define 
\begin{align*}
E_i(\theta) &:=\theta^\top X_i \in I(Y_i,\widehat{Y_i}) \\
C(\theta) &:=\sum\limits_{i=1}^{n_{\rm te}} \mathbbm{1}_{E_i(\theta)}
\end{align*}
Intuitively, $C(\theta)$ is the number of residual intervals containing $\theta^\top X$ across the test examples. Our confidence region is defined as the set of all $\theta$ such that $C(\theta)$ is not abnormally low. \Cref{prop:coverage} uses \Cref{eq:conformal_window} to lower bound $C(\theta_\star)$ in probability. A complete proof is provided in the appendix.

\begin{theorem}
\label{prop:coverage} Under \Cref{assum:cond_inde} and (\Cref{assum:A_a} or \Cref{assum:A_b}), for any $k\leq n_{\rm te}$, it holds
\begin{equation}
\label{eq:coverage_guarantee}
    \mathbb{P}_{\pmb{X_{te}, Y_{te}}}[C(\theta_\star) \geq k] \geq S_{n_{\rm te}} (k, b)
\end{equation}
where
$$ S_{n_{\rm te}} (k, b) = \sum_{j=k}^{n_{\rm te}} \binom{n_{\rm te}}{j}b^j(1-b)^{n_{\rm te}-j} .$$
\end{theorem}
Given $\alpha \in (0,1)$, we define
\begin{equation}
\label{eq:k_alpha}
k_{n_{\rm te}}(\alpha, b) = \max\left\{k \in [n_{\rm te}] : S_{n_{\rm te}} (k, b) \geq 1-\alpha\right\}.
\end{equation}

\Cref{prop:coverage} gives us the tool to finally define a confidence region with finite sample valid coverage guarantees, under our mild assumptions on the noise. From Equations \eqref{eq:coverage_guarantee} and \eqref{eq:k_alpha}, the following proposition immediately follows.
\begin{prop}
For a confidence level $\alpha\in (0,1)$, let us define the confidence region as
\begin{equation}
    \label{eq:confidence_region}
    \Theta_{\alpha}(\pmb{X,Y}) = \left\{\theta\in\mathbb{R}^d : C(\theta)\geq k_{n_{\rm te}}(\alpha, b)\right\}.
\end{equation}
Under the model specification of \Cref{sec:problem_setting}, it holds
$$\mathbb{P}_{\pmb{X_{\rm te},Y_{\rm te}}}[\theta_\star\in \Theta_{\alpha}(\pmb{X,Y})]\geq 1-\alpha.$$
\end{prop}

\begin{rem}
    The probability is taken over only $\pmb{X_{\rm te}, Y_{\rm te}}$ as our guarantee is valid with any $\pmb{\widehat{Y}_{{\rm te}}}$ that verifies \Cref{eq:indep_pred_test}, and thus in particular it is valid for any realization of $\pmb{X_{\rm tr}, Y_{\rm tr}}$, even if $\pmb{\widehat{Y}_{{\rm te}}}$ depends on it.
\end{rem}

\Cref{fig:coverage_by_b} shows the guaranteed coverage $1-\alpha=S_{n_{\rm te}}(k, b)$ from \Cref{eq:coverage_guarantee} as a function of $b$, at fixed $n_{\rm te}=30$ and for $k\in[4,8,12,16]$. For $b=0$ and any $\theta\in\mathbb{R}^d$, any variable $Y$ can be represented as $\theta^\top X + \varepsilon$ using $\varepsilon=Y-\theta^\top X$, which fits the vacuously true \Cref{assum:A_a} with $b=0$. Therefore we can not guarantee a positive coverage unless taking $\mathbb{R}^d$ as confidence region. As $b$ increases, our model becomes increasingly restrictive, which allows the guaranteed coverage of the confidence region to increase, reaching its maximum when the noise is restricted to having a median of $0$ everywhere ($b=0.5$). Increasing $k$ leads to smaller confidence regions (due to the constraints in \Cref{eq:confidence_region} becoming stronger), but also decreases the guaranteed coverage, following \Cref{eq:coverage_guarantee}.

\subsection{Representation as a MILP feasible set}

The expression in \Cref{eq:confidence_region} suffices for efficient membership testing (ie. determining whether a given $\theta$ is in $\Theta_{\alpha}(\pmb{X,Y})$) but is not straightforward to infer, for instance, confidence intervals on specific parameter coordinates. 
We now show how $\Theta_{\alpha}(\pmb{X,Y})$ can be represented as the feasible set of an MILP, which allows optimization of linear objectives for reasonably small $n_{\rm te}$. \looseness=-1

First, let us note from \Cref{eq:confidence_region} that $\theta\in \Theta_{\alpha}$ iff there are at least $k_{n_{\rm te}}(\alpha, b)$ events $E_i$ that are true, and thus iff there exists $(a_i)_{i\in [n_{\rm te}]}\in\{0,1\}^{n_{\rm te}}$ such that
\begin{align}
\label{eq:pre-big-M}
\begin{split}
    &\sum\limits_{i=1}^{n_{\rm te}} a_i\geq k_{n_{\rm te}}(\alpha, b), \\
    &\forall i \in [n_{\rm te}], (a_i = 1) \Longrightarrow E_i.
\end{split}
\end{align}
We can finally use the standard Big-M method \citep{hillier2001introduction, wolsey2014integer} by picking a constant $M$ with a larger order of magnitude than $Y$, and we obtain:
\begin{align}
\begin{split}
    &\sum\limits_{i=1}^{n_{\rm te}} a_i\geq k_{n_{\rm te}}(\alpha, b), \\
    &\forall i,\, \min(Y_i,\widehat{Y_i}) - (1-a_i)M\leq \theta^\top X_i \\
    &\forall i,\, \theta^\top X_i \leq \max(Y_i,\widehat{Y_i}) + (1-a_i)M.\label{eq:bigM}
\end{split}
\end{align}

If $\Theta_{\alpha}$ is bounded and $M$ is sufficiently large, the constraints in \Cref{eq:bigM} will only be active when $a_i=1$, in which case they become equivalent to $\theta^\top X_i\in I(Y_i,\widehat{Y_i})$. It is possible to confirm that the solutions obtained with \Cref{eq:bigM} indeed have inactive constraints when $a_i=0$ (and increase $M$ if needed). With such mechanism in place, the feasible set of \Cref{eq:bigM} is the same as \Cref{eq:pre-big-M}. 

Thus, $\theta\in\Theta_{\alpha}$ iff \eqref{eq:bigM} is satisfied for some binary $(a_i)$. To optimize an objective linear in $\theta$ over $\Theta_{\alpha}$, we can simply introduce the binary slack variables $(a_i)$ and optimize over the set of constraints of \Cref{eq:bigM}, which indeed yields an MILP.

\subsection{Boundedness of $\Theta_{\alpha}$}

Given a set of $k_{n_{\rm te}}(\alpha,b)$ test inputs with a linear span of dimension at most $d-1$,  one can find a direction orthogonal to that span, and displacing $\theta$ in that direction will not affect the corresponding inequalities in \Cref{eq:bigM}. Thus, if $k_{n_{\rm te}}(\alpha,b)<d$, then $\Theta_{\alpha}$ is necessarily empty or not bounded. Conversely, if any subset of $k_{n_{\rm te}}(\alpha,b)$ test inputs has a span of dimension at least $d$, then $\Theta_{\alpha}$ is guaranteed to be bounded, because the solution set will be bounded regardless of which constraints are active. Thus, for applications where a bounded confidence region is desirable or necessary, such as finding confidence intervals on the coordinates, $k_{n_{\rm te}}(\alpha,b)$ should be larger than $d$. Since $k_{n_{\rm te}}(\alpha,b)$ is determined by \Cref{eq:k_alpha}, we can make $k_{n_{\rm te}}(\alpha,b)$ sufficiently large by increasing the test size $n_{\rm te}$, at the cost of an increase in computation complexity.

\section{Applications}
\label{sec:applications}

In this section, we show how the MILP formulation from Section \ref{sec:construction} can be directly leveraged for several applications of interest.

\subsection{Confidence Interval on Coordinates}
\label{subsec:confidence_intervals}

\begin{figure*}[ht]
    \centering
    \includegraphics[trim={4cm 0 4cm 0}, width=\linewidth]{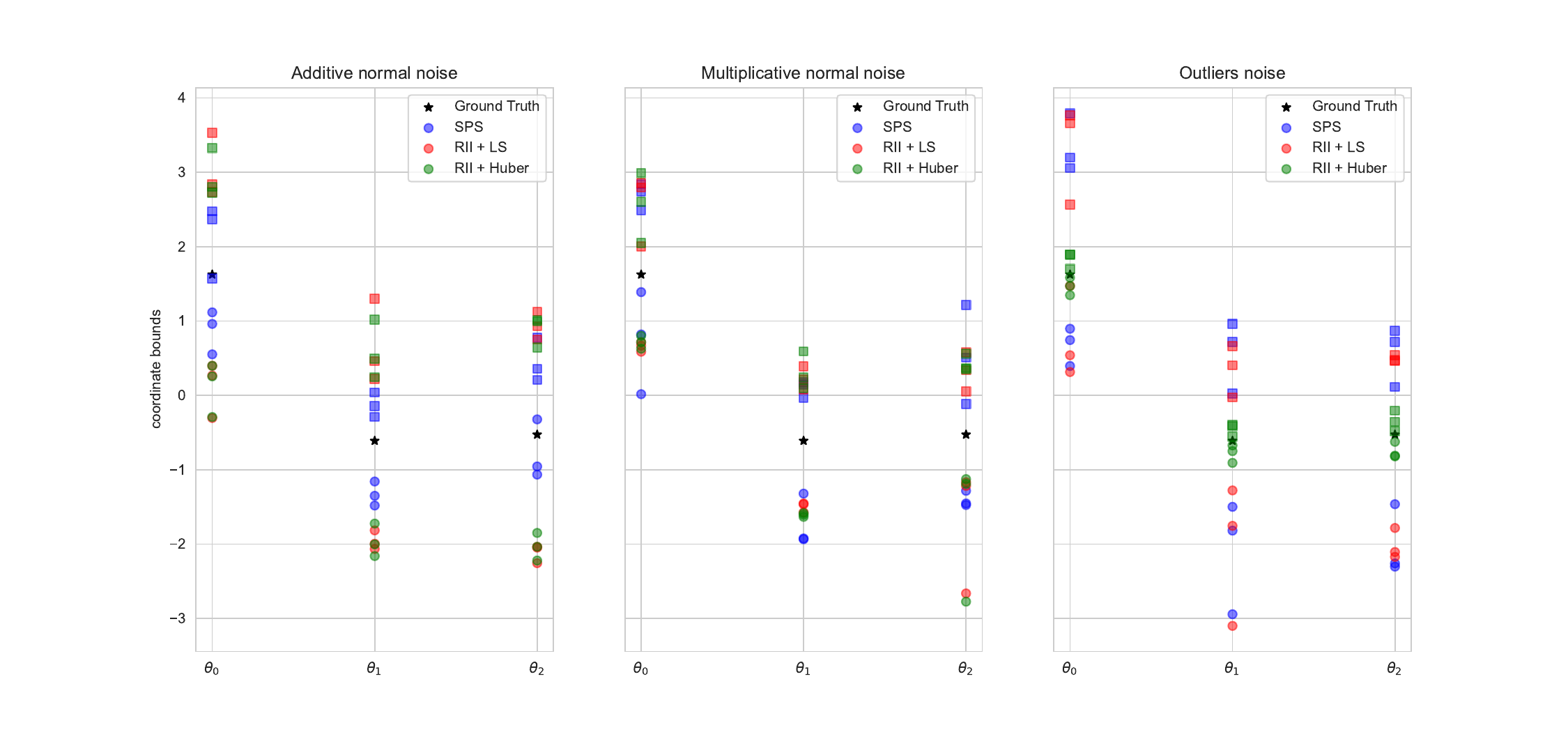}
    \caption{Illustration of the bounds covering the ground-truth parameter $\theta_\star$ under various configurations of noise, for $\alpha=0.1$. Squares correspond to upper bounds while circles denote corresponding lower bounds.}
    \label{fig:bounds}
\end{figure*}

\begin{table*}[t]
\centering

\caption{Coverage of the SPS over-bound and RII across different dimensions and types of noise, for $\alpha=0.1$.}
\label{table:cov}
\vspace{2mm}
\begin{tabular}{cccccccccc}
\hline
           & \multicolumn{3}{c}{Additive Gaussian} & \multicolumn{3}{c}{Multiplicative Gaussian} & \multicolumn{3}{c}{Outliers} \\
           & $d=3$        & $d=10$        & $d=50$       & $d=3$          & $d=10$          & $d=50$         & $d=3$     & $d=10$     & $d=50$    \\ \hline
SPS outer    & 96.8\%     & 100.0\%     & -          & 97.2\%       & 100.0\%       & -            & 98.4\%  & 100.0\%  & -       \\
RII + LS & 89.7\%     & 91.1\%      & 89.1\%     & 91.1\%       & 89.0\%        & 90.1\%       & 89.5\%  & 91.6\%   & 90.8\%  \\ \hline
\end{tabular}
\end{table*}

Perhaps one of the most straightforward application is to deduce confidence intervals on the coordinates of $\theta$. For instance, to compute a lower bound on the $i$-th coordinate of $\theta$ over $\Theta_{\alpha}$, we can solve the following MILP:
\begin{align}
\label{eq:confidence_intervals}
\begin{split}
   &\min\limits_{\theta\in\mathbb{R}^d, a \in \{0,1\}^{n_{\rm te}}}\quad  \theta_i \\
\textrm{s.t.} \quad &\sum\limits_{i=1}^{n_{\rm te}} a_i\geq k_{n_{\rm te}}(\alpha, b), \\
    & \forall i, \min(Y_i,\widehat{Y_i}) - (1-a_i)M-\theta^\top X_i \leq 0 \\
    & \forall i, \theta^\top X_i -  \max(Y_i,\widehat{Y_i}) - (1-a_i)M \leq 0.
\end{split}
\end{align}
For any $\theta\in \Theta_{\alpha}$, all coordinates of $\theta$ are within the confidence intervals calculated by \Cref{eq:confidence_intervals} by construction. Thus with probability at least $1-\alpha$, $\theta_\star\in\Theta_{\alpha}$, which implies all confidence intervals simultaneously contain the ground truth. As a result, every confidence interval contains the corresponding ground truth coordinate with probability at least $1-\alpha$.

These confidence intervals can then be used for interpretability and feature selection. For instance, assuming features have been normalized (or have a similar order of magnitude), a tight confidence interval around $0$ indicates that the corresponding feature likely has low relevance to the output, and may be pruned. Similarly, a very large confidence interval implies that changes in the corresponding coefficient can be compensated with other coefficients, and thus that the feature is likely redundant. \looseness=-1

\subsection{Robust Optimization}

A typical application for confidence regions is robust optimization, in which we seek to find optimal solutions under uncertainty. Robust optimization has been successfully applied in operations research, signal processing, control theory, and other fields.

One of the most important paradigm in robust optimization is Wald's minimax model \citep{wald39, wald45}, which aims at finding the parameters $w$ with best worst-case outcomes over a given uncertainty set for parameter $\theta$. Given our confidence set $\Theta_{\alpha}$ for $\theta$, and assuming (for simplicity) that $W$, the feasible set of $w$, does not depend on $\theta$, we obtain the following optimization problem:
\begin{equation}
   \label{eq:robust_opt}
   \min\limits_{w\in W}\max\limits_{\theta\in \Theta_{\alpha}}\quad f(w, \theta) 
\end{equation}
Where $f(\cdot,\cdot)$ is a cost function to minimize. For instance, $\theta$ could represent the unknown parameters of a dynamical system, and $w$ the controller parameters.

Prior works have explored robust optimization with mixed-integer constraints and convex objective functions \citep{robustOptim}. These methods can be directly applied with $\Theta_{\alpha}$ to perform robust optimization. Alternatively, since MILPs can be difficult to solve, $\Theta_{\alpha}$ can be relaxed to the covering orthotope induced by the confidence intervals of section \ref{subsec:confidence_intervals}, thereby removing the mixed-integer variables and simplifying the resolution.\looseness=-1

\subsection{Hypothesis Testing}

In previous works, confidence regions are typically built with the least square estimator at their center, and may never be empty. On the contrary, it is possible for $\Theta_{\alpha}$ to be empty, which is equivalent to rejecting the null hypothesis $\mathcal{H}_0$ that the data is distributed according to the model described in Section \ref{sec:problem_setting}, with p-value $\alpha$. Indeed, if the null hypothesis is true with parameters $\theta_\star$, then there is probability at least $1-\alpha$ that $\theta_\star\in \Theta_{\alpha}$, and thus $\Theta_{\alpha}$ is empty with probability at most $\alpha$. Notably, $\alpha$ is directly tied to the tolerance level $b$, meaning that we can immediately infer p-values for different values of $b$.

If the predictor $\widehat{Y}$ is linear in $X$, then its coefficients $\hat{\theta}$ belong in $\Theta_{\alpha}$ by construction, and the null hypothesis can not be rejected. Accordingly, to reject the null hypothesis, $\widehat{Y}$ must not be linear in $X$. For instance, $\widehat{Y}$ can be a predictor based on a non-linear model such as XGBoost \citep{Chen_2016}.

Alternatively, $\widehat{Y}$ could be linear in a different variable $Z$, such as a non-linear transformation of $X$. This can provide a framework for feature selection. The null hypothesis is unlikely to be rejected in this setting, unless $\widehat{Y}$ captures the distribution of $Y$ better than any linear function of $X$. This is a powerful result because it applies to \textit{all} linear models of $X$ simultaneously, not just a specific estimator that might be overfitting, and thus indicates that $X$ may inherently lack expressivity.

\section{Experiments}
\label{sec:experiments}

\begin{table*}[t]
\centering
\caption{Rejection rate (ie. frequency at which $\Theta_\alpha$ is empty) of RII with $b=0.5$ for three different functions that are not linear in $X$, and coverage rate when $b$ is set to $\bar{b}$, the largest value of $b$ such that the function falls within our model. $\alpha$ is fixed to $0.1$.}
\label{table:rej}
\vspace{2mm}
\begin{tabular}{cccccc}
\hline
\multicolumn{3}{c}{\begin{tabular}[c]{@{}c@{}}Rejection rate \\ $(b=0.5)$\end{tabular}}                                                                                                         & \multicolumn{3}{c}{\begin{tabular}[c]{@{}c@{}}Coverage rate\\ $(b=\bar{b})$\end{tabular}}                                                                                                    \\ \hline
\begin{tabular}[c]{@{}c@{}}easy \\ $\bar{b}=0.05$\end{tabular} & \begin{tabular}[c]{@{}c@{}}med\\ $\bar{b}=0.14$\end{tabular} & \begin{tabular}[c]{@{}c@{}}hard\\ $\bar{b}=0.27$\end{tabular} & \begin{tabular}[c]{@{}c@{}}easy\\ $\bar{b}=0.05$\end{tabular} & \begin{tabular}[c]{@{}c@{}}med\\ $\bar{b}=0.14$\end{tabular} & \begin{tabular}[c]{@{}c@{}}hard\\ $\bar{b}=0.27$\end{tabular} \\ \hline
$100\%$                                                          & $70\%$                                                         & $4\%$                                                           & $90.4\%$                                                        & $92.6\%$                                                       & $92.4\%$                                                        \\ \hline
\end{tabular}
\end{table*}

\begin{table*}[ht]
\centering
\caption{Average width of the confidence intervals on the coordinates of $\theta$ obtained with SPS, RII with least square estimator, and RII with Huber estimator, for three different types of noise. $\alpha$ is fixed to $0.1$.}
\label{table:width}
\vspace{2mm}
\begin{tabular}{cccc}
\hline
\multirow{2}{*}{} & \multicolumn{3}{c}{noise}                        \\ \cline{2-4} 
                  & Add. normal    & Mult. normal   & Outliers       \\ \hline
SPS               & \textbf{1.230} & 1.903          & 2.633          \\
RII + LS          & 2.861          & \textbf{1.875} & 2.486          \\
RII + Huber       & 2.766          & 1.958          & \textbf{0.363} \\ \hline
\end{tabular}
\end{table*}

We now conduct experiments to evaluate the applications of RII on synthetic data. When applicable, we compare our results with the over-bound of SPS, which to the best of our knowledge is the only prior work constructing finite sample valid confidence regions under weak assumptions on the noise, in a compact form that facilitates applications. Other existing methods typically only allow membership testing, and explicitely delineating the confidence regions is generally intractable.

We evaluate RII using two types of ad hoc predictors: the ordinary least square predictor trained on the training data, and the Huber predictor \citep{10.1214/aoms/1177703732}, which is designed to be robust to outliers, trained on the same training data. 

In all of our experiments, we set $\alpha=0.1$.

For reproducibility, the full settings of our experiments, when not specified in this section, are detailed in \Cref{appendix:exps}.

\paragraph{Linear Toy Examples}
Let $\mathcal{N}(\mu, \sigma)$ the normal distribution of mean $\mu$ and standard deviation $\sigma$.
We sample the ground-truth parameter $\theta_\star$ from a Gaussian distribution, and $X$ is sampled independently from a uniform distribution over $[0,1]^d$. The outputs are generated using three different types of noise
\begin{align*}
(\varepsilon|X) &\sim \mathcal{N}(0,0.5)  & \text{ (additive Gaussian)} \\
(\varepsilon|X) &\sim \mathcal{N}(0,\theta_\star^\top X) & \text{ (multiplicative Gaussian)}\\
(\varepsilon|X) &\sim \mathcal{P}_O & \text{(outliers)}
\end{align*}
$$\text{With } \mathcal{P}_O := B\times\mathcal{N}(0,10)+(1-B)\times\mathcal{N}(0,0.05),$$
where $B$ is a random variable with $\mathbb{P}(B=1)=0.1$ and $\mathbb{P}(B=0)=0.9$.
Intuitively, the noise of type outliers has probability $0.1$ to be sampled from a high variance Gaussian, and probability $0.9$ to be sampled from a low variance Gaussian.

We first evaluate the empirical coverage by sampling independently $\theta_\star, \pmb{X,Y}$. We take a sample size of $1000$ and measure how often $\theta_\star$ falls in the set $\Theta_\alpha(\pmb{X,Y})$, for different types of noise and dimension $d\in[3,10,50]$. The results are reported in \Cref{table:cov}. We do not report the coverage for SPS at $d=50$, as the explicit derivation of the over-bound becomes too computationally expensive at this dimension. This is not a limitation, as SPS also provides a method to test membership very efficiently. However, this fast version only allows membership testing, making it less applicable, so we focus on the over-bound instead. \Cref{table:cov} confirms the 90\% coverage guarantees are indeed respected, with a much tighter coverage in the case of RII than for SPS. 

We then evaluate the coordinate bounds following the process described in \Cref{subsec:confidence_intervals}. In the case of SPS, it requires solving a Quadratic Program (QP), instead of a MILP for RII. We fix $\theta_\star$ with $d=3$ and for each of the three noise types described above, we compute the bounds obtained for three independent samples of $\pmb{X,Y}$. These bounds are presented in \Cref{fig:bounds}, and average width of the confidence intervals in \Cref{table:width}. While SPS performs significantly better in the standard scenario of additive Gaussian noise, all methods perform similarly with multiplicative normal noise, and RII performs better with both predictors in the outliers setting. This indicates that RII might perform more robustly on non-standard noises in comparison to SPS. Additionally, RII with Huber performs significantly better with outliers noise, due to the robustness of the Huber predictor to outliers. This illustrates how the flexibility of RII to use any predictor can bring significant benefits.

\paragraph{Non-Linear Toy Examples}
Here we consider a non-linear feature representation of the input $X$ as $\phi(X) = (X, \sin(8 \pi \|X\|_2)$, along with the linear model\looseness=-1
$$Y = \theta_\star^\top X + v_\star \sin(8\pi\|X\|_2)+ \mathcal{N}(0,1)$$
This model verifies \Cref{assum:A_a} with $b=\bar{b}<0.5$ that depends on $v_\star$. We sample three values of $v_\star$ for which we evaluate the corresponding $\bar{b}$. In \Cref{table:rej} we report the rejection rate using $b=0.5$ (in which case the functions indeed do not verify our model) with $d=3$, and the coverage rate when we use $b=\bar{b}$, ie. when we adjust the tolerance level to include the function within our model. We find that RII is indeed able to reject the null hypothesis when $\bar{b}$ is low, but in the hard example with $\bar{b}=0.27$, the detection rate is not statistically significant as it is below $\alpha=0.1$. This indicates that RII is capable of rejecting the null hypothesis in this setting, but only when the function significantly deviates from linear behavior. When using $b=\bar{b}$, the coverage rate is above $1-\alpha$, as guaranteed by our theoretical results.\looseness=-1

\paragraph{Computational Cost} A significant limitation of RII is that solving an MILP can take exponential time in the worst-case, and become intractable when $n_{\rm te}$ or the $d$ become too large. However, modern solvers can often solve MILP problems in reasonable time. We evaluate and discuss computation times in \Cref{appendix:computation}.\looseness=-1

\section{Conclusion}
\label{sec:conclusion}

Traditionally, a confidence interval is created for a parameter $\theta_\star$ in a statistical model like \Cref{eq:linear_model} by first obtaining an estimate $\hat \theta$ from the data. The next step involves examining the distribution of estimation errors $\hat \theta - \theta_\star$ to construct the interval by thresholding at some appopriate quantile level. However, a closed form distribution is notoriously difficult to obtain using most methods. As such, the standard guarantees are obtained upon asymptotic normality, as is the case when using the maximum likelihood principle to obtain the estimator. In contrast, RII rely on inverting the confidence set constructed around the prediction $\theta_\star^\top X$, which enables us to bypass the majority of preceding assumptions. Consequently, we can operate using any predictors and under less stringent assumptions on the noise. RII is a flexible method, capable of levering any ad hoc predictor, of rejecting the linearity of the observed data, and displaying promising robustness across noise distributions. 

\looseness=-1


\bibliographystyle{plainnat}
\bibliography{arxiv}

\begin{thebibliography}{20}
\providecommand{\natexlab}[1]{#1}
\providecommand{\url}[1]{\texttt{#1}}
\expandafter\ifx\csname urlstyle\endcsname\relax
  \providecommand{\doi}[1]{doi: #1}\else
  \providecommand{\doi}{doi: \begingroup \urlstyle{rm}\Url}\fi

\bibitem[Angelopoulos et~al.(2023)Angelopoulos, Bates, Fannjiang, Jordan, and
  Zrnic]{angelopoulos2023prediction}
Anastasios~N Angelopoulos, Stephen Bates, Clara Fannjiang, Michael~I. Jordan,
  and Tijana Zrnic.
\newblock Prediction-powered inference.
\newblock \emph{arXiv preprint arXiv:2301.09633}, 2023.

\bibitem[Campi and Weyer(2005)]{CAMPI20051751}
M.C. Campi and E.~Weyer.
\newblock Guaranteed non-asymptotic confidence regions in system
  identification.
\newblock \emph{Automatica}, 41:\penalty0 1751--1764, 2005.

\bibitem[Chen and Guestrin(2016)]{Chen_2016}
Tianqi Chen and Carlos Guestrin.
\newblock {XGBoost}.
\newblock In \emph{Proceedings of the 22nd {ACM} {SIGKDD} International
  Conference on Knowledge Discovery and Data Mining}. {ACM}, 2016.

\bibitem[Csaji et~al.(2015)Csaji, Campi, and Weyer]{Csaji_2015}
Balazs~Csanad Csaji, Marco~Claudio Campi, and Erik Weyer.
\newblock Sign-perturbed sums: A new system identification approach for
  constructing exact non-asymptotic confidence regions in linear regression
  models.
\newblock \emph{{IEEE} Transactions on Signal Processing}, 63, jan 2015.

\bibitem[Csáji et~al.(2012)Csáji, Campi, and Weyer]{CSAJI2012227}
Balázs~Csanád Csáji, Marco~C. Campi, and Erik Weyer.
\newblock Non-asymptotic confidence regions for the least-squares estimate.
\newblock \emph{IFAC Proceedings Volumes}, 45, 2012.

\bibitem[Dalai et~al.(2007)Dalai, Weyer, and Campi]{DALAI20071418}
Marco Dalai, Erik Weyer, and Marco~C. Campi.
\newblock Parameter identification for nonlinear systems: Guaranteed confidence
  regions through lscr.
\newblock \emph{Automatica}, 43, 2007.

\bibitem[Daniels(1954)]{10.1214/aoms/1177728718}
H.~E. Daniels.
\newblock {A Distribution-Free Test for Regression Parameters}.
\newblock \emph{The Annals of Mathematical Statistics}, 25, 1954.

\bibitem[{den Dekker} et~al.(2008){den Dekker}, Bombois, and {Van den
  Hof}]{DENDEKKER20085024}
Arnold~J. {den Dekker}, Xavier Bombois, and Paul~M.J. {Van den Hof}.
\newblock Finite sample confidence regions for parameters in prediction error
  identification using output error models.
\newblock \emph{IFAC Proceedings Volumes}, 41, 2008.

\bibitem[Dobriban and Lin(2023)]{dobriban2023joint}
Edgar Dobriban and Zhanran Lin.
\newblock Joint coverage regions: Simultaneous confidence and prediction sets,
  2023.

\bibitem[Draper and Smith(1981)]{draper1981applied}
N.R. Draper and H.~Smith.
\newblock \emph{Applied Regression Analysis}.
\newblock Wiley, 1981.

\bibitem[Gevers(2006)]{4019326}
M.~Gevers.
\newblock A personal view of the development of system identification: A
  30-year journey through an exciting field.
\newblock \emph{IEEE Control Systems Magazine}, 26, 2006.

\bibitem[Hillier and Lieberman(2001)]{hillier2001introduction}
F.S. Hillier and G.J. Lieberman.
\newblock \emph{Introduction to Operations Research}.
\newblock McGraw-Hill, 2001.

\bibitem[Huber(1964)]{10.1214/aoms/1177703732}
Peter~J. Huber.
\newblock {Robust Estimation of a Location Parameter}.
\newblock \emph{The Annals of Mathematical Statistics}, 35, 1964.

\bibitem[Jochmans(2022)]{doi:10.1080/01621459.2020.1831924}
Koen Jochmans.
\newblock Heteroscedasticity-robust inference in linear regression models with
  many covariates.
\newblock \emph{Journal of the American Statistical Association}, 117, 2022.

\bibitem[Senov et~al.(2014)Senov, Amelin, Amelina, and Granichin]{MSPS}
Alexander Senov, K.~Amelin, Natalia Amelina, and O.~Granichin.
\newblock Exact confidence regions for linear regression parameter under
  external arbitrary noise.
\newblock \emph{Proceedings of the American Control Conference}, 2014.

\bibitem[Siddiqui et~al.(2015)Siddiqui, Gabriel, and Azarm]{robustOptim}
Sauleh Siddiqui, Steven Gabriel, and Shapour Azarm.
\newblock Solving mixed-integer robust optimization problems with interval
  uncertainty using benders decomposition.
\newblock \emph{Journal of the Operational Research Society}, 66, 2015.

\bibitem[Wald(1939)]{wald39}
Abraham Wald.
\newblock Contributions to the theory of statistical estimation and testing
  hypotheses.
\newblock \emph{The Annals of Mathematical Statistics}, 10, 1939.

\bibitem[Wald(1945)]{wald45}
Abraham Wald.
\newblock Statistical decision functions which minimize the maximum risk.
\newblock \emph{Annals of Mathematics}, 46, 1945.

\bibitem[Wasserman et~al.(2020)Wasserman, Ramdas, and
  Balakrishnan]{doi:10.1073/pnas.1922664117}
Larry Wasserman, Aaditya Ramdas, and Sivaraman Balakrishnan.
\newblock Universal inference.
\newblock \emph{Proceedings of the National Academy of Sciences}, 117:\penalty0
  16880--16890, 2020.

\bibitem[Wolsey and Nemhauser(2014)]{wolsey2014integer}
L.A. Wolsey and G.L. Nemhauser.
\newblock \emph{Integer and Combinatorial Optimization}.
\newblock Wiley, 2014.

\end{thebibliography}

\newpage
\onecolumn
\appendix

\section{Proofs}
    \label{appendix:proofs}

\subsection{Lemma 1}
Under \Cref{assum:cond_inde} and (\Cref{assum:A_a} or \Cref{assum:A_b}), for any test point $(X, Y)$ with prediction $\widehat{Y}$, it holds
\begin{equation}
\label{eq:appendix:conformal_window}
\mathbb{P}_{X, Y}\bigg(\theta_{\star}^{\top} X\in I(Y,\widehat{Y})\bigg) \geq b .
\end{equation}

\textbf{Proof.} Let $x\in dom(X)$ and $\widehat{Y}\in\mathbb{R}$. If $\widehat{Y}\geq \theta_\star^\top x$, then 
$$\varepsilon(x) \leq 0 \Longrightarrow  \theta_\star^\top x + \varepsilon \leq \theta_\star^\top x \leq \widehat{Y} \Rightarrow \theta_\star^\top x \in I(\theta_\star^\top x + \varepsilon, \widehat{Y}),$$
and thus $\mathbb{P}_{Y|X=x}\left(\theta_\star^\top x \in I\left(Y,\widehat{Y}\right)\right)\geq \mathbb{P}_\varepsilon(\varepsilon\leq 0)$.

Similarly, if $\widehat{Y}\leq \theta_\star^\top x$, then 
$$\varepsilon(x) \geq 0 \Longrightarrow \theta_\star^\top x + \varepsilon \geq \theta_\star^\top x \geq \widehat{Y} \Rightarrow \theta_\star^\top x \in I(\widehat{Y}, \theta_\star^\top x + \varepsilon), $$
and thus $\mathbb{P}_{Y|X=x}\left(\theta_\star^\top x \in I\left(Y,\widehat{Y}\right)\right)\geq \mathbb{P}_\varepsilon(\varepsilon\geq 0)$.

Therefore, $\forall x\in dom(X), \widehat{Y}\in\mathbb{R}$, $$\mathbb{P}_{Y|X=x}\left(\theta_\star^\top x\in I\left(Y,\widehat{Y}\right)\right)\geq \min(\mathbb{P}_\varepsilon(\varepsilon\leq 0\mid x),\, \mathbb{P}_\varepsilon(\varepsilon\geq 0\mid x))=d_\varepsilon(x).$$
Let us first assume \Cref{assum:cond_inde} and \Cref{assum:A_a}.
We recall that from \Cref{assum:A_a},
\begin{equation} 
    \label{eq:appendix:b-bounded_quantile-1}
    \mathbb{E}_X[d_\varepsilon(X)] \geq b.
\end{equation}
Finally, 
%
%
\begin{align*}
\mathbb{P}_{X,Y}\left(\theta_\star^\top X \in I\left(Y,\widehat{Y}\right)\right)&=\int \mathbb{P}_{Y|X=x}\left(\theta_\star^\top x\in I\left(Y,\widehat{Y}\right)\right) p(x) dx \\
&\geq\int d_\varepsilon(x) p(x) dx\geq b
\end{align*}
from \Cref{eq:appendix:b-bounded_quantile-1}, where $p(\cdot)$ represents the pdf of $X$.

Let us now assume \Cref{assum:cond_inde} and \Cref{assum:A_b}. We have 
\begin{align*}
\mathbb{P}_{X,Y}\left(\theta_\star^\top X \in I\left(Y,\widehat{Y}\right)\right) &\geq \min_{x}\mathbb{P}_{Y|X=x}\left(\theta_\star^\top x\in I\left(Y,\widehat{Y}\right)\right)\\
&\geq \min_{x}(d_\varepsilon(x))\geq b 
\end{align*}
from \Cref{assum:A_b}, which concludes the proof.

\subsection{Theorem 1}
Under \Cref{assum:cond_inde} and (\Cref{assum:A_a} or \Cref{assum:A_b}), for any $k\leq n_{\rm te}$, it holds
\begin{equation}
    \mathbb{P}_{\pmb{X_{te}, Y_{te}}}[C(\theta_\star) \geq k] \geq S_{n_{\rm te}} (k, b)
\end{equation}

\textbf{Proof.} Let $E_i$ the events defined as in \Cref{sec:construction}. 

Under \Cref{assum:cond_inde} and \Cref{assum:A_a}, the $X_i$ are sampled independently and the $\varepsilon_i$ are sampled independently given $X_i$. Thus, $\mathbb{P}(E_i|E_1,\dots,E_{i-1})=\mathbb{P}(E_i)\geq b$ (from \Cref{prop:conformal_window}).

Under \Cref{assum:cond_inde} and \Cref{assum:A_b}, we have 
$$\mathbb{P}(E_i|E_1,\dots,E_{i-1})\geq \min_x\mathbb{P}_{Y_i|X_i=x}(E_i|E_1,\dots,E_{i-1})=\min_x\mathbb{P}_{Y_i|X_i=x}(E_i)\geq b$$ (the equality comes from \Cref{assum:cond_inde} and the last inequality from the proof of \Cref{prop:conformal_window} above). 

Whether \Cref{assum:A_a} or \Cref{assum:A_b} is verified, we obtain $\mathbb{P}(E_i|E_1,\dots,E_{i-1})\geq b$.

If $n_{\rm te}=1$, then it is easy to verify that for any $k\leq n_{\rm te}$, $\mathbb{P}(\sum\limits_{i=1}^{n_{\rm te}}\mathbbm{1}_{E_i}\geq k)\geq\sum\limits_{i=k}^{n_{\rm te}}{n_{\rm te} \choose i}b^i(1-b)^{n_{\rm te}-i}$.

Let us assume that up to some $n_{\rm te}\in\mathbb{N}$, $\forall k\leq n_{\rm te}$, $\mathbb{P}(\sum\limits_{i=1}^{n_{\rm te}}\mathbbm{1}_{E_i}\geq k)\geq\sum\limits_{i=k}^{n_{\rm te}}{n_{\rm te} \choose i}b^i(1-b)^{n_{\rm te}-i}$.

We now consider $n_{\rm te}+1$ trials.
Then for $k=n_{\rm te}+1$, 
\begin{align*}
\mathbb{P}(\sum_{i=k}^{n_{te}+1}\mathbbm{1}_{E_i}\geq k) &=\mathbb{P}(\forall i\leq n_{\rm te}+1, E_i)\\
&=\prod\limits_{i=1}^{n_{\rm te}+1}\mathbb{P}(E_i|E_1,\dots,E_{i-1})\geq b^{n_{\rm te}+1}\\
&=\sum_{i=k}^{n_{\rm te}+1}{n_{\rm te}+1 \choose i}b^i(1-b)^{n_{\rm te}+1-i}.
\end{align*}

For $k\leq n_{\rm te}$, at least $k$ of $E_1,\dots, E_{n_{\rm te}+1}$ are true iif at least $k$ of $E_0,\dots,E_{n_{\rm te}}$ are true and $E_{n_{\rm te}+1}$ is false or at least $k-1$ of $E_1,\dots,E_{n_{\rm te}}$ are true and $E_{n_{\rm te}+1}$ is true. Thus, 
\begin{equation}
\begin{aligned}
\mathrm{cov} &=\mathbb{P}\left(\sum_{i=1}^{n_{\rm te}+1}\mathbbm{1}_{E_i}\geq k \right) \\
&=\mathbb{P}(\sum\limits_{i=1}^{n_{\rm te}}\mathbbm{1}_{E_i}\geq k)\times(1-\mathbb{P}(E_{n_{\rm te}+1}|E_1,\dots,E_{n_{\rm te}}))
+\mathbb{P}(\sum\limits_{i=1}^{n_{\rm te}}\mathbbm{1}_{E_i}\geq k-1)\times\mathbb{P}(E_{n_{\rm te}+1}|E_1,\dots,E_{n_{\rm te}})\\
&\geq\sum\limits_{i=k}^{n_{\rm te}}{n_{\rm te} \choose i}b^i(1-b)^{n_{\rm te}-i}\times(1-\mathbb{P}(E_{n_{\rm te}+1}|E_1,\dots,E_{n_{\rm te}}))
+\sum\limits_{i=k-1}^{n_{\rm te}}{n_{\rm te} \choose i}b^i(1-b)^{n_{\rm te}-i}\times\mathbb{P}(E_{n_{\rm te}+1}|E_1,\dots,E_{n_{\rm te}})\\
&\geq \sum\limits_{i=k}^{n_{\rm te}}{n_{\rm te} \choose i}b^i(1-b)^{n_{\rm te}-i}
+{n_{\rm te} \choose k-1}b^{k-1}(1-b)^{n_{\rm te}-k+1}\times\mathbb{P}(E_{n_{\rm te}+1}|E_1,\dots,E_{n_{\rm te}})\\
&\geq \sum\limits_{i=k}^{n_{\rm te}}{n_{\rm te} \choose i}b^i(1-b)^{n_{\rm te}-i} + {n_{\rm te} \choose k-1}b^{k-1}(1-b)^{n_{\rm te}-k+1}\times b\\
&\geq \sum\limits_{i=k}^{n_{\rm te}}{n_{\rm te} \choose i}b^i(1-b)^{n_{\rm te}-i}\times(1-b)
+\sum\limits_{i=k-1}^{n_{\rm te}}{n_{\rm te} \choose i}b^i(1-b)^{n_{\rm te}-i}\times b\\
&\geq \sum\limits_{i=k}^{n_{\rm te}}{n_{\rm te} \choose i}b^i(1-b)^{n_{\rm te}-i+1}+\sum\limits_{i=k-1}^{n_{\rm te}}{n_{\rm te} \choose i}b^{i+1}(1-b)^{n_{\rm te}-i}\\
&\geq \sum\limits_{i=k}^{n_{\rm te}+1}{n_{\rm te} \choose i}b^i(1-b)^{n_{\rm te}-i+1}+\sum\limits_{i=k}^{n_{\rm te}+1}{n_{\rm te} \choose i-1}b^{i}(1-b)^{n_{\rm te}-i+1}\\
&\geq \sum\limits_{i=k}^{n_{\rm te}+1}\left[{n_{\rm te} \choose i}+{n_{\rm te} \choose i-1}\right] b^i(1-b)^{n_{\rm te}-i+1}\\
&\geq \sum\limits_{i=k}^{n_{\rm te}+1}{n_{\rm te}+1 \choose i}b^i(1-b)^{n_{\rm te}+1-i}
\end{aligned}
\end{equation}

By induction, we can thus conclude that for any $n_{\rm te}$ and $k\leq n_{\rm te}$, 
$$\mathbb{P}(C(\theta_\star)\geq k):=\mathbb{P}\left(\sum\limits_{i=k}^{n_{\rm te}}\mathbbm{1}_{E_i}\geq k\right)\geq \sum\limits_{i=k}^{n_{\rm te}}{n_{\rm te} \choose i}b^i(1-b)^{n_{\rm te}-i}=S_{n_{\rm te}}(k, b).$$ 

\subsection{Proposition 1}
For a confidence level $\alpha\in (0,1)$, let us define the confidence region as
\begin{equation}
    \Theta_{\alpha}(\pmb{X,Y}) = \left\{\theta\in\mathbb{R}^d : C(\theta)\geq k_{n_{\rm te}}(\alpha, b)\right\}.
\end{equation}
Under the model specification of \Cref{sec:problem_setting}, it holds
$$\mathbb{P}_{\pmb{X_{\rm te},Y_{\rm te}}}[\theta_\star\in \Theta_{\alpha}(\pmb{X,Y})]\geq 1-\alpha.$$

\textbf{Proof.} The proof immediately follows from Equations \eqref{eq:coverage_guarantee} and \eqref{eq:k_alpha}. Indeed, we have

$$\mathbb{P}_{\pmb{X_{\rm te},Y_{\rm te}}}[\theta_\star\in \Theta_{\alpha}(\pmb{X,Y})]=\mathbb{P}_{\pmb{X_{\rm te},Y_{\rm te}}}[C(\theta_\star)\geq k_{n_{\rm te}}(\alpha, b)]\geq S_{n_{\rm te}}(k_{n_{\rm te}}(\alpha,b))\geq 1-\alpha.$$

\section{Experimental details}
    \label{appendix:exps}

For all RII experiment requiring to solve the MILP, we use a $M$ constant (for the relaxation) of $50$. We find no occurrences where more than $k$ inequalities are active, which indicates there is no need to increase $M$. Such occurences would occur if $k<d$ however, as the confidence region would then not be bounded.

\subsection{Linear Toy Examples}


To measure coverage in \Cref{table:cov}, $\theta_\star$ is fixed for \Cref{fig:bounds}, while it is resampled for each trial of \Cref{table:cov} used to estimate the coverage. Each coordinate of $\theta_\star$ is sampled independently from $\mathcal{N}(0,1)$. The noise is sampled as described in \Cref{sec:experiments}. We use $60$ training points. For SPS, we use $q=1$ and $m=10$. We found that using $q=10$ and $m=100$ led to nearly identical performances, but at a higher cost. The coordinates of $x$ are sampled independently from the uniform distribution on $[0,1]$. For RII, we use $k=16$ and $n_{\text{te}}=39$, which leads to $\alpha(k, n_{\text{te}}, b)\approx 0.0998\leq 0.1$.

\begin{table}[h]
    \centering
    \begin{tabular}{lccc}
        \toprule
        Scenario & Parameters & Resampling & Rate Parameters \\
        \midrule
        \Cref{fig:bounds} & $d=3$ & Fixed $\theta_\star$ & - \\
        \Cref{table:cov} & $d=3, 10, 50$ & Resampled $\theta_\star$ & $k=16$, $n_{\text{te}}=39$, $\alpha \approx 0.1$ \\
        \bottomrule
    \end{tabular}
    \caption{Experimental setup details for \Cref{fig:bounds} and \Cref{table:cov}.}
    \label{table:exp_setup}
\end{table}

\subsection{Non-Linear Toy Examples}


For \Cref{table:rej} we employ three different values of $v_\star$, with the functional form introduced in \Cref{sec:experiments}. We run $100$ trials for each function and measure the frequency at which the confidence region is empty with $k=16$ and $n_{\text{te}}=39$ for the rejection rate, which corresponds to $\alpha(k, n_{\text{te}}, b=0.5)\approx 0.1$.

\begin{table}[h]
    \centering
    \begin{tabular}{lccc}
        \toprule
        Example & $\bar{b}$ & $v_\star$ & Parameters for Rate \\
        \midrule
        "Easy" & 0.05 & 0.05 & $k=2$, $n_{\text{te}}=74$ \\
        "Med"  & 0.14 & 0.2  & $k=7$, $n_{\text{te}}=73$ \\
        "Hard" & 0.27 & 0.1  & $k=10$, $n_{\text{te}}=50$ \\
        \bottomrule
    \end{tabular}
    \caption{Values of $\bar{b}$ and $v_\star$ for different examples.}
    \label{table:rej}
\end{table}

For the coverage rate, we use $k=10$ and $n_{\text{te}}=50$ for the hard example, $k=7$ and $n_{\text{te}}=73$ for the med example, and $k=2$ with $n_{\text{te}}=74$ for the easy example. All these pairs correspond to values $\alpha(k, n_{\text{te}}, \bar{b})\leq 0.1$.

For $\widehat{Y}$, we generate predictions using ordinary least squares on $(x, \sin(8\pi\|x\|_2))$.

\section{Computational cost}
    \label{appendix:computation}

MILP problems are NP-hard, so the worst-case computation cost of solving over our confidence region is exponential. However, modern solvers often manage to solve such problems much faster in average, while still obtaining certifiably optimal solutions.
In comparison, SPS doesn't have discrete variable, but possesses quadratic constraints, leading to a quadratically constrained linear program. Moreover, finding the radius of the ellipsoid of SPS requires solving $m$ convex semidefinite programs.  

RII can perform membership testing in very short time (it only requires to evaluate $n_{\rm te}$ linear inequalities, no optimization is required, and similarly for SPS (when using the exact form and not the over-bound). We report in \Cref{table:time} the computation time required to instantiate the confidence region (e.g. compute the ellipsoid axes and radius or train the estimator) and the computation time required to optimize a single linear objective once instantiated, as the wall clock time on a personal laptop. We use PULP\_CBC as a solver, which is under an eclipse v2.0 license. 

RII is substantially more costly for inferring a single linear objective, and the method is difficult to scale to large dimensions (ie $d>50$ approximately, a problem also shared by SPS). However, this computation cost remains accessible for problems of reasonable dimension, despite using $39$ discrete variables in this example. It may be possible to approximate the MILP used to optimize linear objectives with RII to reduce the computational burden and scale to larger dimensions, which we leave to future works.

\begin{table*}[ht]
\centering
\caption{Average wall clock time on a personal laptop to instantiate the confidence region and to optimize a single linear objective for respectively SPS and RII, at $d=3$ with $k=16$ and $n_{\rm te}=39$.}
\label{table:time}
\vspace{2mm}
\begin{tabular}{lll}
\hline
    & Conf. Region Instantiation & Linear Obj. Optim \\ \hline
SPS &             0.0270s               &        0.0013s          \\
RII &               0.0007s             &        2.7972s           \\ \hline
\end{tabular}
\end{table*}

\end{document}